\documentclass[11pt,a4paper]{article}
\usepackage[hyperref]{emnlp2020}
\usepackage{times}
\usepackage{latexsym}

\usepackage{microtype}

\aclfinalcopy 
\def\aclpaperid{6} 

\usepackage{booktabs}
\usepackage{float}
\usepackage{subfigure}
\usepackage{tabularx}
\usepackage{wasysym}

\usepackage{makecell} 
\usepackage{multirow} 

\title{Determining Question-Answer Plausibility in Crowdsourced Datasets Using Multi-Task Learning}

\author{Rachel Gardner\hspace{5mm}Maya Varma\hspace{5mm}Clare Zhu\hspace{5mm}Ranjay Krishna\\
  Department of Computer Science, Stanford University, CA \\
  \texttt{\{rachel0, mvarma2, clarezhu, ranjaykrishna\}@cs.stanford.edu} \\
  }

\date{2020}

\begin{document}
\maketitle

\begin{abstract}
Datasets extracted from social networks and online forums are often prone to the pitfalls of natural language, namely the presence of unstructured and noisy data. In this work, we seek to enable the collection of high-quality question-answer datasets from social media by proposing a novel task for automated quality analysis and data cleaning: \textit{question-answer (QA) plausibility}. Given a machine or user-generated question and a crowd-sourced response from a social media user, we determine if the question and response are valid; if so, we identify the answer within the free-form response.
 
We design BERT-based models to perform the QA plausibility task, and we evaluate the ability of our models to generate a clean, usable question-answer dataset. Our highest-performing approach consists of a single-task model which determines the plausibility of the question, followed by a multi-task model which evaluates the plausibility of the response as well as extracts answers (Question Plausibility AUROC=0.75, Response Plausibility AUROC=0.78, Answer Extraction F1=0.665).
\end{abstract}

\section{Introduction}
Large, densely-labeled datasets are a critical requirement for the creation of effective supervised learning models. The pressing need for high quantities of labeled data has led many researchers to collect data from social media platforms and online forums \cite{abuelhaija2016youtube8m, Thomee_2016, sentiment140}. Due to the presence of noise and the lack of structure that exist in these data sources, manual quality analysis (usually performed by paid crowdworkers) is necessary to extract structured labels, filter irrelevant examples, standardize language, and perform other preprocessing tasks before the data can be used. However, obtaining dataset annotations in this manner is a time-consuming and expensive process that is often prone to errors. 

In this work, we develop automated data cleaning and verification mechanisms for extracting high-quality data from social media platforms\footnote{All code is available at \url{https://github.com/rachel-1/qa_plausibility}.}. We specifically focus on the creation of question-answer datasets, in which each data instance consists of a question about a topic and the corresponding answer. In order to filter noise and improve data quality,  we propose the task of \textit{question-answer (QA) plausibility}, which includes the following three steps:
\begin{itemize}
\item \textit{Determine question plausibility:} Depending on the type of dataset being constructed, the question posed to respondents may be generated by a machine or a human. We determine the likelihood that the question is both relevant and answerable. 
\item \textit{Determine response plausibility:} We predict whether the user's response contains a reasonable answer to the question. 
\item \textit{Extract answer from free-form response:} If the response is deemed to be plausible, we identify and extract the segment of the response that directly answers the question. 
\end{itemize}

Because we assume social media users generally answer questions in good faith (and are posed questions which they can answer), we can assume plausible answers are correct ones \cite{hcomp}. Necessarily, if this property were not satisfied, then any adequate solutions would require the very domain knowledge of interest. Therefore, we look to apply this approach toward data with this property.

In this study, we demonstrate an application of QA plausibility in the context of visual question answering (VQA), a well-studied problem in the field of computer vision \cite{vqa}. We assemble a large VQA dataset with images collected from an image-sharing social network, machine-generated questions related to the content of the image, and responses from social media users. We then train a multitask BERT-based model and evaluate the ability of the model to perform the three subtasks associated with QA plausibility. The methods presented in this work hold potential for reducing the need for manual quality analysis of crowdsourced data as well as enabling the use of question-answer data from unstructured environments such as social media platforms.

\section{Related Work}

Prior studies on the automated labeling task for datasets derived from social media typically focus on the generation of noisy labels; models trained on such datasets often rely on weak supervision to learn relevant patterns. However, approaches for noisy label generation, such as Snorkel \cite{snorkel} and CurriculumNet \cite{curriculumnet}, often use functions or other heuristics to generate labels. One such example is the Sentiment140 dataset, which consists of 1.6 million tweets labeled with corresponding sentiments based on the emojis present in the tweet \cite{sentiment140}. In this case, the presence of just three category labels (positive, neutral, negative) simplifies the labeling task and reduces the effects of incorrect labels on trained models; however, this problem becomes increasingly more complex and difficult to automate as the number of annotation categories increases.

Previous researchers have studied question relevance by reasoning explicitly about the information available to answer the question. Several VQA studies have explicitly extracted premises, or assumptions made by questions, to determine if the original question is relevant to the provided image \cite{promise_of_premise, question_relevance}. A number of machine comprehension models have been devised to determine the answerability of a question given a passage of text \cite{squad2, Back2020NeurQuRINQ}. In contrast, we are able to leverage the user's freeform response to determine if the original question was valid. Our model is also tasked with supporting machine-generated questions, which may be unanswerable and lead to noisy user-generated responses.

While the concept of answer plausibility in user responses has also been previously explored, existing approaches use hand-crafted rules and knowledge sources \cite{smith2005}. By using a learned approach, we give our system the flexibility to adapt with the data and cover a wider variety of cases.

\section{Dataset}
\label{sec:Problem Formulation}
\begin{figure}[H]
    \centering
    \includegraphics[scale=0.7]{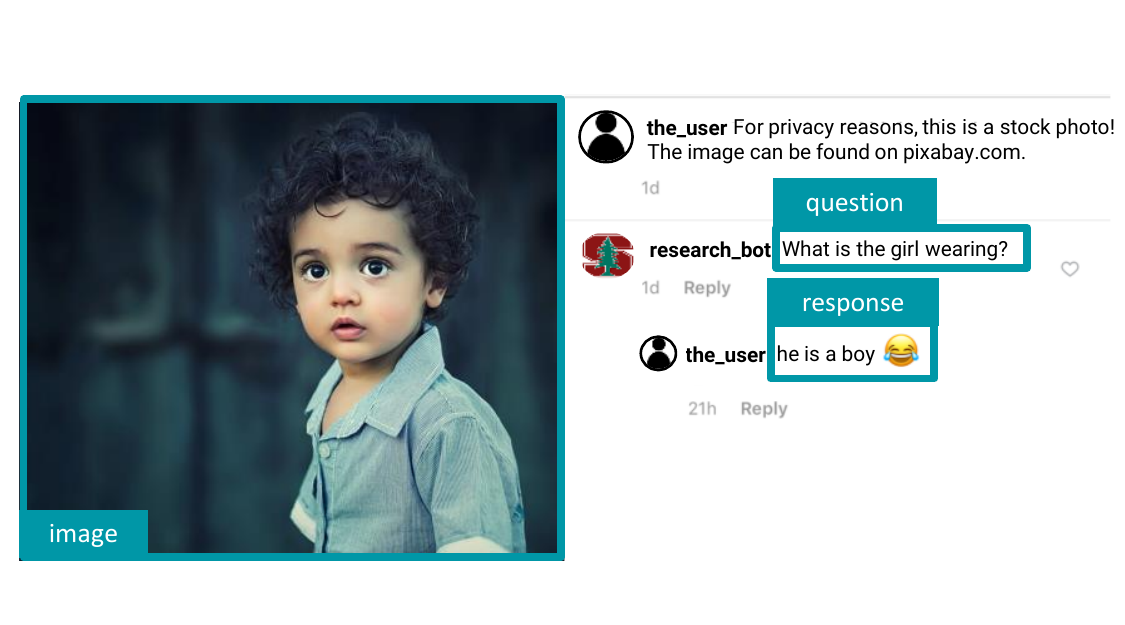}
    \caption{\textit{An example question and response pair collected from social media.} Note that since the questions are generated by a bot, the question may not always be relevant to the image, as demonstrated here.}
    \label{fig:example}
\end{figure}

The dataset consists of questions and responses collected from an image-sharing social media platform. We utilize an automated question-generation bot in order to access public image posts, generate a question based on image features, and record data from users that replied to the question, as shown in \autoref{fig:example} \cite{qbot}. Because the question-generation bot was designed to maximize information gain, it generates questions across a wide variety of categories, including objects, attributes, spatial relationships, and activities (among others). For the sake of space, we refer readers to the original paper for more information on the method of question generation and diversity of the resulting questions asked. All users that contributed to the construction of this dataset were informed that they were participating in a research study, and IRB approval was obtained for this work. For the privacy of our users, the dataset will not be released at this time. Rather than focus on the specific dataset, we wish to instead present a general method for cleaning user-generated datasets and argue its generality even to tasks such as visual-question-answering.

\begin{table}[h]
    \small
    \begin{center}
        \begin{tabular}{|c|c|c|c|c|}
            \Xhline{2\arrayrulewidth}
            \# & \multicolumn{2}{c}{Example Question/Response} & Plausible? & \%\\
            \Xhline{2\arrayrulewidth}
            \multirow{2}{*}{1.} & Q & What is on the table? & Y & \multirow{2}{*}{50.6} \\\cline{2-4}
            & R & beet and carrot juice\smiley{}\smiley{} & Y & \\
            \Xhline{2\arrayrulewidth}
            \multirow{2}{*}{2.} & Q & What is the person doing? & Y & \multirow{2}{*}{22.8} \\\cline{2-4}
            & R & not much lol &  N & \\
            \Xhline{2\arrayrulewidth}
            \multirow{2}{*}{3.} & Q & What is on top of the cake? &  N & \multirow{2}{*}{11.4} \\\cline{2-4}
            & R & that is not cake that's chicken & Y & \\
            \Xhline{2\arrayrulewidth}
            \multirow{2}{*}{4.} & Q & What is the hamster doing? &  N & \multirow{2}{*}{15.3} \\\cline{2-4}
            & R & that is not a hamster & N & \\
            \Xhline{2\arrayrulewidth}
            \end{tabular}
    \end{center}
\caption{\textit{Representative examples of cases present in the data, and the percentage of examples represented by each class in our dataset.} Examples (1) and (2) have valid questions that accurately refer to the corresponding images, while (3) and (4) do not correctly refer to objects in the image. However, in example (3), the user identifies the error made by the bot and correctly refers to the  object in the image; as a result, this response is classified as valid.}  

\label{table:qualitative-eval-no-image-needed}
\end{table}

\begin{figure*}[ht]
    \centering
    \includegraphics[scale=0.6]{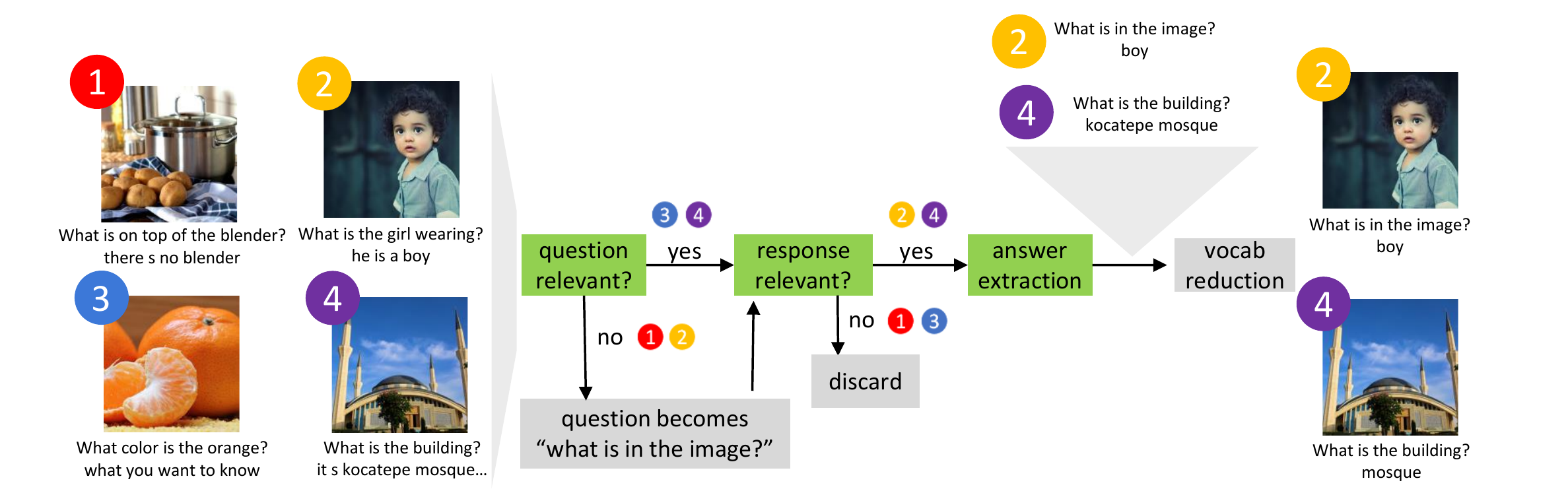}
    \caption{\textit{Overview of the QA plausibility task, with representative examples}. Given a question and user response, we determine if the question and response are plausible given the image. If so, we then extract a structured answer label from the response.}
    \label{fig:pipeline_example}
\end{figure*}

The dataset was labeled by crowdworkers on Amazon Mechanical Turk (AMT), who performed three annotation tasks, as shown in \autoref{table:qualitative-eval-no-image-needed}: (1) determine if the question was plausible, (2) determine if the response was plausible, and (3) if the response was deemed to be plausible, extract an answer span. Plausible questions and answers are defined as those that accurately refer to the content of the image. 
 
It is important to note that since the question-generation process is automated, the question could be unrelated to the image due to bot errors; however, in such situations where the question is deemed to be implausible, the response may still be valid if it accurately refers to the content of the image. If the response is judged to be plausible, the AMT crowdworker must then extract the answer span from the user’s response. In order to capture the level of detail we required (while discouraging AMT crowdworkers from simply copy/pasting the entire response), we set the maximum length of an answer span to be five words for the labeling step. However, the final model itself is not limited to answers of any particular length.

For cost reasons, each example was labeled by only one annotator. While we could have averaged labels across annotators, we found that the majority of the labeling errors were due to misunderstandings of the non-standard task, meaning that errors were localized to particular annotators rather than randomly spread across examples. This issue was mitigated by adding a qualifying task and manually reviewing a subset of labels per worker for the final data collection.

While one might expect images to be necessary (or at least helpful) for determining question and response plausibility, we found that human annotators were able to determine the validity of the inputs based solely on text without the need for the accompanying image. In our manual analysis of several hundred examples (approximately 5\% of the dataset), we found that every example which required the image to label properly could be categorized as a ``where'' question. When the bot asked questions of the general form ``where is the X'' or ``where was this taken,'' users assumed our bot had basic visual knowledge and was therefore asking a question not already answered by the image (such as ``where is the dog now" or ``what part of the world was this photo taken in"). This led to valid responses that did not pertain to image features and were therefore not helpful for training downstream models. \autoref{table:qualitative-eval-image-needed} gives one such example. Once we removed these questions from the dataset, we could not find a single remaining example that required image data to label properly. As a result, we were able to explore the QA plausibility task in a VQA setting, despite not examining image features.

\begin{table}[h]
    \small
    \begin{center}
        \begin{tabular}{|c|c|c|c|}
            \Xhline{2\arrayrulewidth}
            \multirow{3}{*}{\includegraphics[scale=0.2]{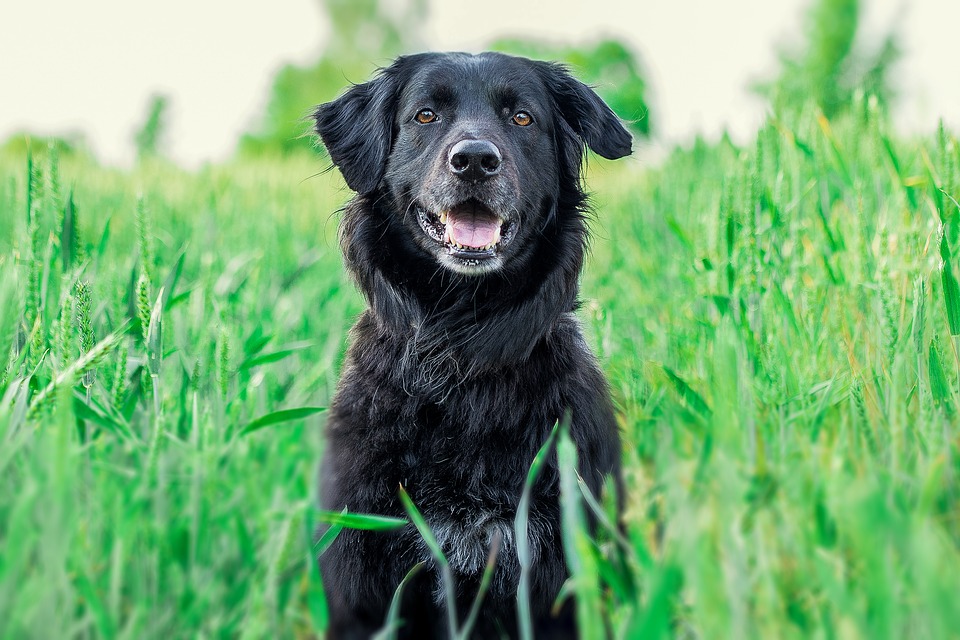}} & \multicolumn{2}{c}{Example Question/Response} & Valid?\\\cline{2-4}
            & Q & Where is the dog? & Y\\\cline{2-4}
            & R & sitting next to me on the sofa & N\\
            \Xhline{2\arrayrulewidth}
        \end{tabular}
    \end{center}
\caption{\textit{Example requiring analysis of the original image (removed from dataset along with other ``where" questions which often lead to confusion)}.}
\label{table:qualitative-eval-image-needed}
\end{table}

Our preprocessing steps and annotation procedure resulted in a total of 7200 question-response pairs with answer labels. We use a standard split of 80\% of the dataset for training, 10\% for validation, and 10\% for testing. 

\section{Models and Experiments}
\begin{figure}[H]
    \centering
    \includegraphics[scale=0.5]{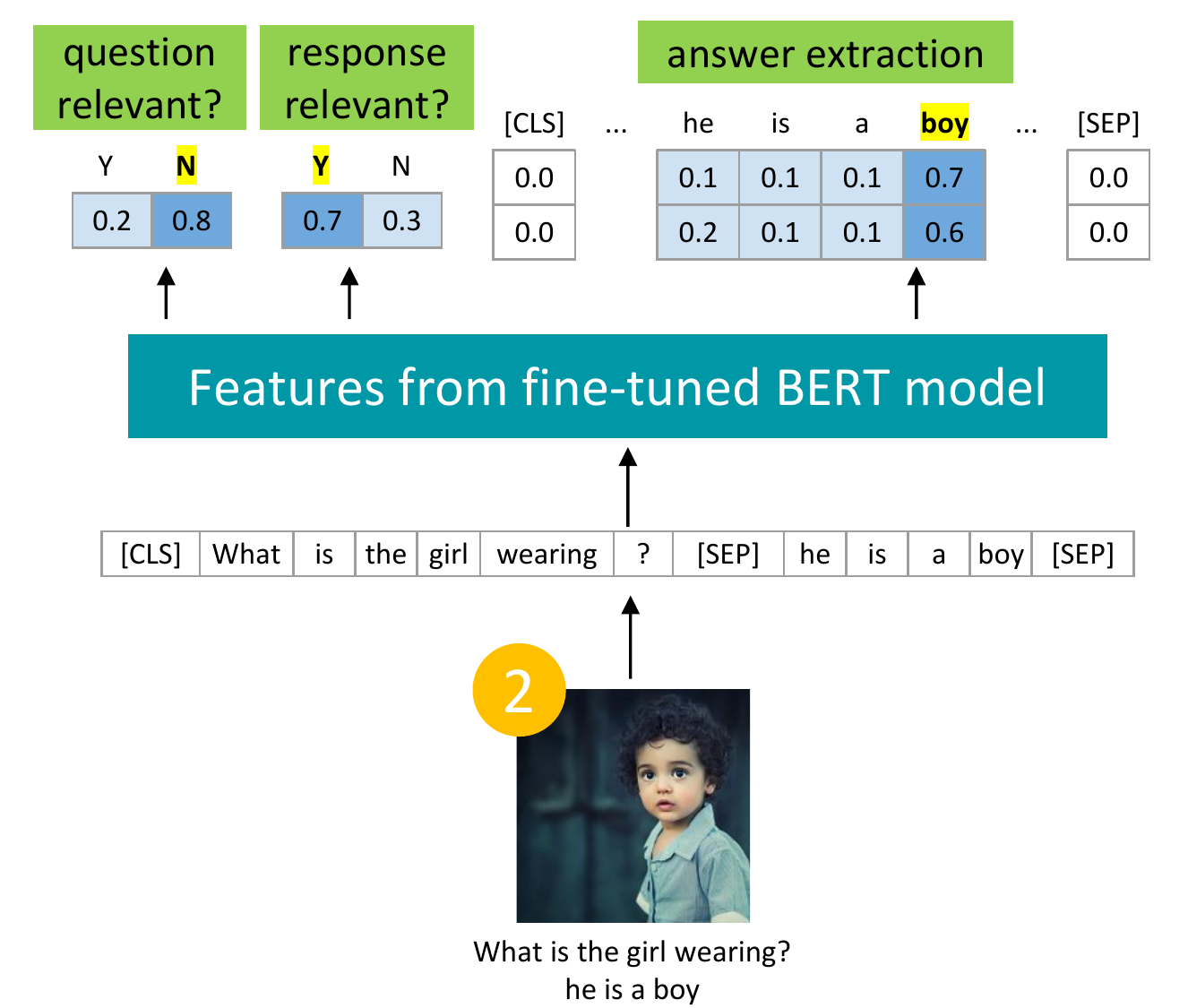}
    \caption{\textit{Model architecture.} The question and user response serve as input to a modified BERT model, which will output question plausibility, response plausibility, and an answer label.}
    \label{fig:bert_model}
\end{figure}

\textbf{Model Architecture:} As shown in \autoref{fig:bert_model}, we utilized a modified BERT model to perform the three sub-tasks associated with QA plausibility. The model accepts a concatenation of the machine-generated question and user response as input, with the [CLS] token inserted at the start of the sentence and the [SEP] token inserted to separate the question and response.

\begin{table*}[ht]
    \begin{center}
        \begin{tabular}{|c|cc|cc|c|}
        \toprule
        Combined Task &  \multicolumn{2}{|c|}{Question Plausibility}  &  \multicolumn{2}{|c|}{Response Plausibility} & \multicolumn{1}{|c|}{Answer Extraction} \\
        {} & Acc & AUROC & Acc & AUROC & F1 \\
        \midrule
        Question Plausibility (QP) only & \textbf{65.51}\% & \textbf{0.7488} & - & - & - \\
        Response Plausibility (RP) only & - & - & 64.62\% & 0.7674 & - \\
        Answer Extraction only & - & - & - & - & 0.568 \\
        RP and Answer Extraction & - & - & \textbf{70.13\%}	& \textbf{0.7870} & \textbf{0.665}  \\
        QP, RP and Answer Extraction & 63.90\%	& 0.6803 & 60.91\% & 0.6881 & 0.6160 \\
        \bottomrule
        \end{tabular}
    \end{center}
\caption{\textit{Model Evaluation Metrics.} Performance metrics of our model are shown here. Multi-task learning helps improve performance when the model is simultaneously trained on the response plausibility and answer extraction subtasks, but decreases performance when the model is simultaneously trained on all three subtasks. }
\label{table:multi_task}
\end{table*} 

In order to perform the question plausibility classification task, the pooled transformer output is passed through a dropout layer (p=0.5), fully connected layer, and a softmax activation function. An identical approach is used for response plausibility classification. To extract the answer span, encoded hidden states corresponding to the last attention block are passed through a single fully connected layer and softmax activation; this yields two probability distributions over tokens, with the first representing the start token and the second representing the end token. The final model output includes the probability that the question and response are plausible, with each expressed as a score between 0 and 1; if the response is deemed to be plausible, the model also provides the answer label, which is expressed as a substring of the user response. 

\noindent
\textbf{Experiments:} We utilized a pretrained BERT Base Uncased model, which has 12 layers, 110 million parameters, a hidden layer size of 768, and a vocabulary size of 30,522. We trained several single-task and multi-task variants of our model in order to measure performance on the three subtasks associated with QA plausibility. In the multi-task setting, loss values from the separate tasks are combined; however, an exception to this exists if the user’s response is classified as implausible. In these cases, the answer span extraction loss is manually set to zero and the answer extraction head is not updated. 

We evaluated performance on question and response plausibilities by computing accuracy and AUC-ROC scores. Performance on the answer span extraction task was evaluated with F1 scores, which measure overlap between the predicted answer label and the true answer \cite{squad2}. 

\section{Results}
We investigated performance of our BERT model on the various subtasks associated with QA plausibility. Results are summarized in \autoref{table:multi_task}. Single-task models trained individually on the subtasks achieved an AUC-ROC score of 0.75 on the question plausibility task, an AUC-ROC score of 0.77 on the response plausibility task, and an F1 score of 0.568 on the answer extraction task. A multi-task model trained simultaneously on all three tasks demonstrated decreased performance on the question and response plausibility tasks when compared to the single-task models. We found that the highest performance was achieved when a single-task model trained on the question plausibility task was followed by a multi-task model trained on both the response plausibility and answer extraction tasks; this model achieved an AUC-ROC score of 0.75 on question plausibility, an AUC-ROC score of 0.79 on response plausibility, and an F1 score of 0.665 on answer extraction. 

Our results suggest that multi-task learning is most effective when the tasks are closely related, such as with response plausibility and answer extraction. Since the BERT architecture is extremely quick for both training and evaluation, we found that the increase in performance afforded by using a single-task model and multi-task model in series was worth the overhead of training two separate models. It is worth noting that a more complicated model architecture might have been able to better accommodate the loss terms from all three subtasks, but we leave such efforts to future work.  

\section{Discussion}
Deep learning studies are often hindered by lack of access to large datasets with accurate labels. In this paper, we introduced the question-answer plausibility task in an effort to automate the data cleaning process for question-answer datasets collected from social media. We then presented a multi-task deep learning model based on BERT, which accurately identified the plausibility of machine-generated questions and user responses as well as extracted structured answer labels. Although we specifically focused on the visual question answering problem in this paper, we expect that our results will be useful for other question-answer scenarios, such as in settings where questions are user-generated or images are not available. 
   
Overall, our approach can help improve the deep learning workflow by processing and cleaning the noisy and unstructured natural language text available on social media platforms. Ultimately, our work can enable the generation of large-scale, high-quality datasets for artificial intelligence models. 

\bibliographystyle{acl_natbib}
\bibliography{anthology,emnlp2020}

\end{document}